\documentclass[a4paper, 10pt, conference]{ieeeconf}      
\IEEEoverridecommandlockouts                              

\usepackage{cite}
\usepackage{amsmath,amssymb,amsfonts}
\usepackage{algorithmic}
\usepackage{graphicx}
\usepackage{textcomp}
\usepackage{xcolor}
\def\BibTeX{{\rm B\kern-.05em{\sc i\kern-.025em b}\kern-.08em
    T\kern-.1667em\lower.7ex\hbox{E}\kern-.125emX}}

\usepackage{wasysym} 
\usepackage{booktabs}
\usepackage{dirtree}
\usepackage{array}
\usepackage{subcaption}
\usepackage{amsmath}
\usepackage{amssymb}
\usepackage{hyperref}

\title{ATLAS: An \underline{A}nnotation \underline{T}ool for \underline{L}ong-horizon Robotic \underline{A}ction \underline{S}egmentation}

\author{ Sergej Stanovcic$^{1,*}$, Daniel Sliwowski$^{1,*}$, and Dongheui Lee$^{1,2}$
\thanks{$^{*}$ equal contributions.}%
\thanks{$^{1}$Sergej Stanovcic, Daniel Sliwowski and Dongheui Lee are affiliated with the Autonomous Systems Lab, Technische Universität Wien (TU Wien), Vienna, Austria (e-mail: \texttt{\{sergej.stanovcic, daniel.sliwowski, dongheui.lee\}@tuwien.ac.at}).}%
\thanks{$^{2}$Dongheui Lee is also affiliated with the Institute of Robotics and Mechatronics, German Aerospace Center (DLR), Wessling, Germany.}%
\thanks{This work was supported by the European Union project INVERSE under grant agreement No. 101136067 and in part supported by the Robot Industry Core Technology Development Program under Grant No. 00416440 funded by the Korea Ministry of Trade, Industry and Energy (MOTIE).}
\thanks{The annotation tool can be found at \url{https://github.com/TUWIEN-ASL/ATLAS-tuwienasl}.}%
}

\newcolumntype{C}[1]{>{\centering\arraybackslash}m{#1}}

\begin{document}
\maketitle

\begin{abstract}
Annotating long-horizon robotic demonstrations with precise temporal action boundaries is crucial for training and evaluating action segmentation and manipulation policy learning methods. Existing annotation tools, however, are often limited: they are designed primarily for vision-only data, do not natively support synchronized visualization of robot-specific time-series signals (e.g., gripper state or force/torque), or require substantial effort to adapt to different dataset formats. In this paper, we introduce \textsc{ATLAS}, an annotation tool tailored for long-horizon robotic action segmentation. \textsc{ATLAS} provides time-synchronized visualization of multi-modal robotic data, including multi-view video and proprioceptive signals, and supports annotation of action boundaries, action labels, and task outcomes. The tool natively handles widely used robotics dataset formats such as ROS bags and the Reinforcement Learning Dataset (RLDS) format~\cite{ramos2021rlds}, and provides direct support for specific datasets such as \textbf{REASSEMBLE}~\cite{Sliwowski-RSS-25}. \textsc{ATLAS} can be easily extended to new formats via a modular dataset abstraction layer.  Its keyboard-centric interface minimizes annotation effort and improves efficiency. In experiments on a contact-rich assembly task, \textsc{ATLAS} reduced the average per-action annotation time by at least 6\% compared to ELAN~\cite{wittenburg2006elan}, while the inclusion of time-series data improved temporal alignment with expert annotations by more than 2.8\% and decreased boundary error fivefold compared to vision-only annotation tools.
\end{abstract}


\section{Introduction}

Annotated data plays a central role in many machine learning pipelines, as it is required either for model training, evaluation, or both. In supervised learning, annotations are directly used during training to compute the loss. While unsupervised methods do not require annotations at training time, in cases such as unsupervised action segmentation, they still rely on annotated data for quantitative evaluation~\cite{li2024otas}. As a result, the development and evaluation of both supervised and unsupervised models ultimately depends on the availability of annotated datasets. Creating such annotations typically requires dedicated software that allows human annotators to mark relevant information. Over the years, many tools have been developed for computer vision tasks such as object detection, semantic segmentation, and object tracking, including CVAT~\cite{cvat} and LabelMe~\cite{wada2018labelme}. However, these tools do not support the annotation of temporal action boundaries or action outcomes. Action-level annotations are particularly important in robotics, where long-horizon tasks can be decomposed into a sequence of simpler and shorter actions. Prior work has shown that explicit task decomposition is beneficial for learning long-horizon robotic policies~\cite{sun2025arch}, and that decomposed demonstrations enable robots to learn high-level plans for complex tasks~\cite{TAS_robotics}. This motivates the need for tools that support precise temporal action annotation in robotic datasets.

\begin{figure}
    \centering
    \includegraphics[width=\linewidth]{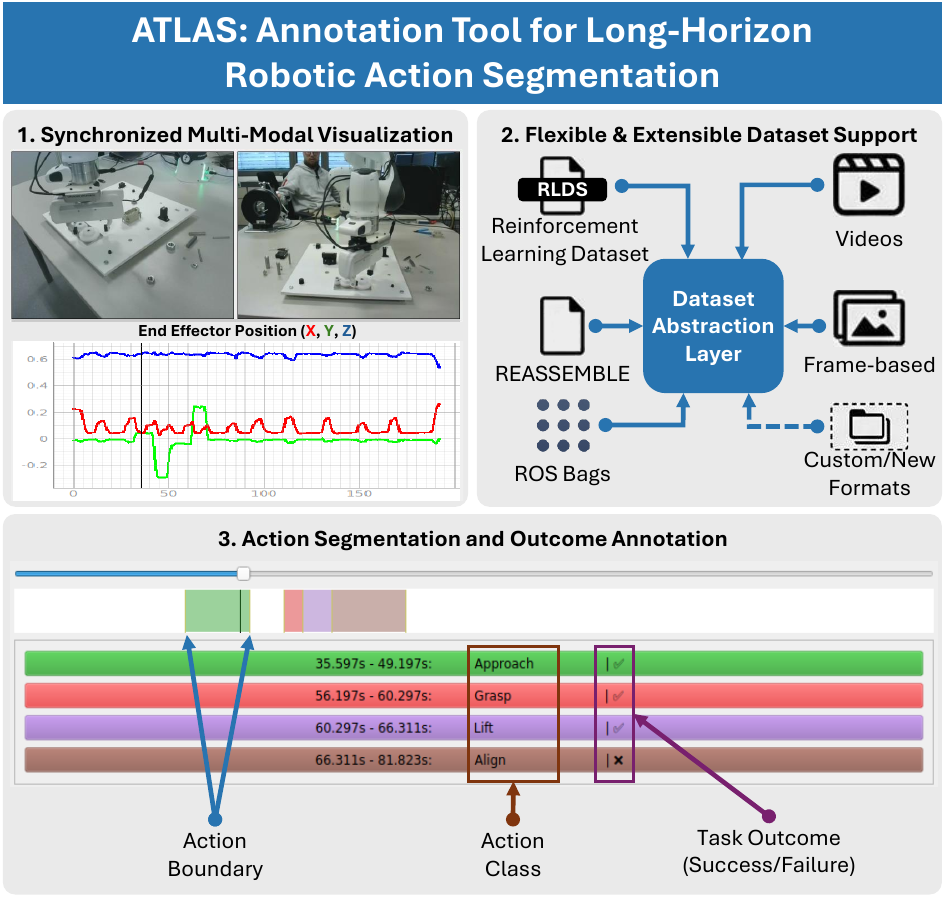}
    \caption{\textsc{ATLAS} is a tool for long-horizon robotic action segmentation, supporting time-synchronized visualization of multi-modal data. It handles common robotics dataset formats and can be extended to new ones (dashed line). The tool allows annotation of action boundaries, classes, and outcomes (success or failure).}

    \label{fig:placeholder}
\end{figure}

\begin{table*}[t]
    \begin{center}
    \caption{Comparison of existing annotation tools. ($\checkmark$ = Supported,  $\times$ = Not Supported)}
    \label{tab:comparison}
    \begin{tabular}{lccc C{5cm}}
    \toprule
    & & \textbf{Time-series} & \textbf{Temporal} & \textbf{Supported Dataset} \\
    \textbf{Application} & \textbf{Domain} & \textbf{Visualization} & \textbf{Segmentation Annotation} & \textbf{Formats} \\
    \midrule
    CVAT \cite{cvat} & Computer Vision & $\times$ & $\times$ & Frames, Videos \\
    Foxglove Studio \cite{foxglove2024} & Robotics & $\checkmark$ & $\times$ & MCAP~\cite{mcap2024}, ROS Bags \\
    ROSAnnotator \cite{zhang2025rosannotator} & Robotics & $\times$\textsuperscript{1} & $\checkmark$ & ROS Bags \\
    Anvil \cite{kipp2001Anvil} & Behavioral & $\checkmark$ & $\checkmark$ & Videos, Motion Capture \\
    ELAN \cite{wittenburg2006elan} & Linguistics & $\checkmark$ & $\checkmark$ & Videos, CSV files \\
    \midrule
    \textbf{Ours} & Robotics & \textbf{$\checkmark$} & \textbf{$\checkmark$} & Frames, Videos, ROS Bags, RLDS, REASSEMBLE \\
    \bottomrule
    \end{tabular}    
    \end{center}
    
    \textsuperscript{1} The authors provide guidelines on how to modify the code to add support; however, this functionality is not implemented in the default version.
\end{table*}

Several existing tools, such as ELAN~\cite{wittenburg2006elan} and Anvil~\cite{kipp2001Anvil}, allow for annotating actions and their temporal boundaries and are widely used in behavioral and linguistic analysis. However, these tools are designed primarily for human-centered data and typically support video, audio, and motion capture visualization. They lack visualization of robot-specific signals such as end-effector pose, gripper state, or force and torque measurements, which are often critical for accurately identifying action boundaries, as not all events are clearly observable from vision alone~\cite{sliwowski2025m2r2}. Other tools, such as Foxglove Studio~\cite{foxglove2024}, provide visualization capabilities for a wide range of robotic data types. However, they are intended for data inspection and do not support annotation creation. The closest existing annotation tool to our approach is ROSAnnotator~\cite{zhang2025rosannotator}, which allows for annotating ROS bag files for human-robot interaction. While it supports camera visualization and audio transcription, it does not directly handle additional robotic data modalities. Extending support for such data requires manual code modifications, as noted by the authors.

An additional challenge arises from the diversity of dataset formats used in robotics and related fields. Existing datasets cover a wide range of representations, including raw video files, ROS bag recordings, task-specific formats such as REASSEMBLE~\cite{Sliwowski-RSS-25}, and more general data schemas such as the Reinforcement Learning Datasets (RLDS) format~\cite{ramos2021rlds}, which is commonly used in robotics. 
Supporting multiple dataset formats, or allowing users to easily add custom dataset formats to the tool, enables the annotation of a broad range of already existing datasets. Adding additional annotations to existing datasets is particularly important for robotics, where collecting new data is expensive and time-consuming, and where valuable datasets already exist but may lack detailed annotations.

In this work, we propose \textsc{ATLAS}, an annotation tool for long-horizon robotic action segmentation. The design of \textsc{ATLAS} is guided by several key criteria motivated by the practical challenges of annotating robotic datasets. First, the tool must support the visualization of time-synchronized data from multiple robotic sensors. This includes, but is not limited to, multi-view camera streams, gripper state, and end-effector pose, which together provide critical context for accurately identifying action boundaries and outcomes. Second, the tool is designed to natively support a wide range of existing dataset formats. In its current version, \textsc{ATLAS} supports raw videos, frame-based datasets, ROS bag files, the REASSEMBLE format, and the Reinforcement Learning Datasets (RLDS) format. Support for RLDS is particularly important, as it enables \textsc{ATLAS} to be applied to many existing robotics datasets without conversion. For example, all datasets in the Open X-Embodiment repository~\cite{o2024open} are distributed in RLDS format, which currently includes 71 datasets comprising a total of 2{,}419{,}193 episodes. Third, \textsc{ATLAS} provides an abstract interface for integrating new datasets, allowing new data formats to be added with minimal effort. This abstraction enables the tool to scale beyond the formats currently supported and facilitates annotation of future datasets. Finally, the tool is designed to minimize annotation time by reducing interactions with the mouse and maximizing keyboard-based control. \textsc{ATLAS} supports customizable keyboard shortcuts, allowing users to adapt the interface to their annotation workflow and improve efficiency.

In summary, \textsc{ATLAS} provides the following capabilities:
\begin{itemize}
    \item Visualization of time-synchronized, multi-modal robotic data, including multi-view RGB videos, gripper state, and end-effector pose.
    \item Native support for multiple dataset formats, including videos, frame-based datasets, ROS bag files, and RLDS, as well as specific robotic datasets such as REASSEMBLE.
    \item An extensible dataset abstraction that enables easy integration of new data formats without modifying the visualization and annotation layer.
    \item Efficient annotation of long-horizon robotic action segments through a keyboard-centric interface with customizable shortcuts.
    \item Support for annotating action boundaries, action labels, and task outcomes (success or failure).
\end{itemize}

\section{Related Work} 
We broadly categorize existing tools for temporal data annotation and visualization into three groups based on their primary design focus: (i) general computer vision annotation tools, (ii) behavioral and linguistic analysis tools, and (iii) robotics-oriented visualization and annotation tools. Table~\ref{tab:comparison} summarizes the capabilities of representative tools from each category and compares them with our proposed system.

\subsection{General Computer Vision Annotation} 
Tools such as CVAT \cite{cvat} and LabelMe \cite{wada2018labelme} are industry standards for object detection and semantic segmentation. They support annotation of spatial information, such as bounding boxes, polygons, and segmentation masks. However, as shown in Table \ref{tab:comparison}, they lack support for synchronized time-series visualization, which is important for robotics applications. While both tools allow video-based annotation and object tracking, they do not support temporal action annotation.

\subsection{Behavioral and Linguistic Analysis} 
Software developed for behavioral and linguistic research, such as ELAN~\cite{wittenburg2006elan} and Anvil~\cite{kipp2001Anvil}, supports temporal action segmentation annotation and the visualization of videos alongside additional data streams. Anvil supports visualization of 3D motion capture data and 2D time-series signals of limb movements aligned with video streams. ELAN, in contrast, allows visualization of arbitrary 2D time-series data; however, such data must be first converted into CSV files and synchronized with the video timeline. In contrast, \textsc{ATLAS} natively supports commonly used robotic dataset formats, like RLDS and ROS Bags, enabling direct visualization and annotation without requiring dataset conversion.

\subsection{Robotics-oriented Annotation Tools} 
Platforms such as Foxglove Studio~\cite{foxglove2024} provide environments for inspecting robot data, including 3D visualizations and real-time plotting. However, they are designed for live data monitoring rather than dataset annotation and do not provide workflows to define, edit, or save temporal action boundaries. As a result, researchers must use external scripts to track and save labels.

ROSAnnotator~\cite{zhang2025rosannotator} integrates ROS bag visualization with temporal labeling capabilities. By default, it supports single-view video and audio transcript visualization, but does not handle additional sensor modalities, such as joint positions, gripper states, or force/torque data. Many annotation actions, such as starting and ending an action segment and selecting its class, rely on mouse interactions, which increases the time required to label each episode. In contrast, \textsc{ATLAS} is designed for keyboard-driven operation, reducing the time annotators spend switching between mouse and keyboard. 

Although ROS bag files are commonly used in robotics, ROSAnnotator does not support other dataset formats, like RLDS, which are commonly used in robotic manipulation learning. \textsc{ATLAS} addresses this limitation through a template method design pattern~\cite{designpatterns1995}: an abstract dataset class defines a common API that specialized dataset classes inherit and implement. This decouples the visualization and annotation frontend from the underlying dataset, allowing new datasets and formats to be integrated with minimal effort.

\section{ATLAS Functions}
\subsection{Importing Data}
To use ATLAS, the user first specifies a configuration file containing basic information about the dataset, available robotic sensor data, cameras, and the path where annotations should be saved. The user can also choose which sensors are displayed by default in the visualization tool and customize the keys used for annotation.
Once ATLAS is started, users can import data by clicking the \textit{Load Data} button, which opens a navigation window for selecting files or directories. After loading, the first episode is displayed in the interface, and annotation can begin.

\subsection{Data Visualization}
The interface of \textsc{ATLAS} is organized into five vertically stacked sections, as illustrated in Figure~\ref{fig:layout}. The uppermost section displays recorded camera data. If multiple cameras are specified in the configuration file, all are shown simultaneously. Beneath the camera view is the \textit{Data Selector}, a scrollable menu listing the time-series data streams (e.g., robot proprioception) specified in the configuration file. Users can select which streams to visualize in the subsequent panel. The next section displays the selected time-series data using PyQtGraph~\cite{pyqtgraph}. All time series plots are interactive and allow zooming and axis range adjustment for detailed inspection. Below the time-series visualization, a textual description of the current episode is displayed. This is particularly useful for datasets such as REASSEMBLE or datasets from the Open X-Embodiment repository (RLDS), where episodes may include pre-annotated natural language descriptions of the episode. If no description is available, the field is left blank. The following section contains a timeline showing the current position within the episode and current annotations. Users can scrub through the video using the mouse on the timeline. Next is the \textit{Annotation Panel}, which lists all action annotations for the episode. Each entry shows the start and end times relative to the episode start, the action class, and the action outcome (success indicated with a checkmark, failure with a cross). Users can modify annotations directly in this panel, adjusting the start and end times, the action label, and the success/failure status. Finally, the bottom section contains controls that can be used to annotate the data. If there exists a keyboard shortcut for the button it is displayed next to the name.

\subsection{Annotation}
\begin{figure*}[t]
  \centering
  \begin{minipage}[c]{0.65\textwidth}
    \centering
    \includegraphics[width=\textwidth]{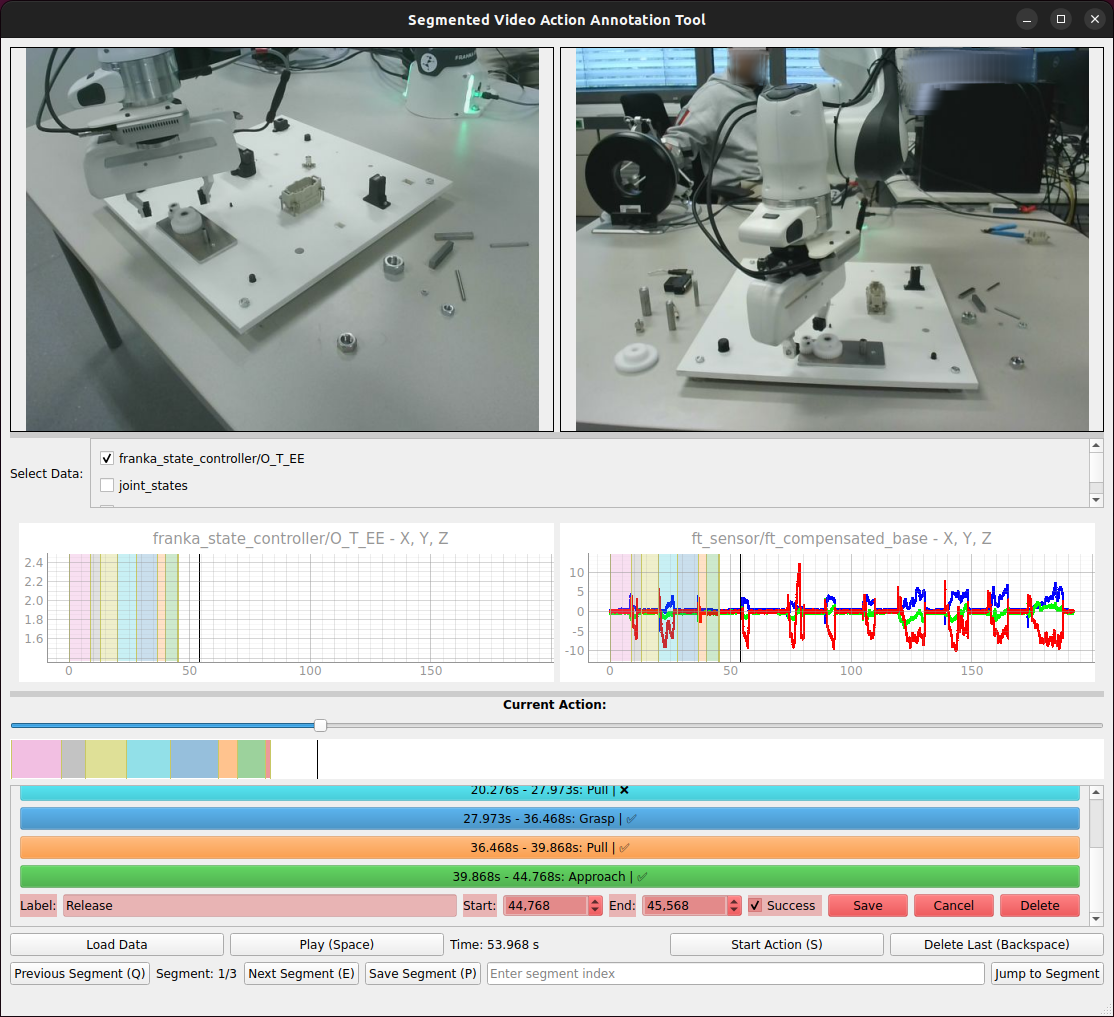}
    \subcaption{}
    \label{fig:layout}
  \end{minipage}%
  \hfill
  \begin{minipage}[c]{0.29\textwidth}
    \includegraphics[width=0.64\textwidth]{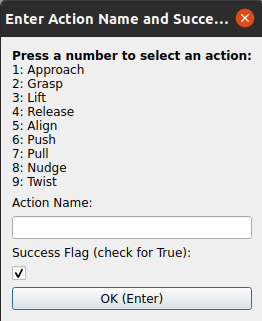}
    \subcaption{}
    \label{fig:action_dialog}
    
    \vspace{1.5em}
    \hspace{-3em}
    \begin{minipage}{\linewidth}
      \begin{minipage}{0.7\linewidth}
        \raggedright
\dirtree{%
    .1 /segmentation\_gui/.
    .2 annotations/.
    .2 configuration/.
    .2 src/.
    .3 segmentation\_gui/.
    .4 datasets/.
    .5 dataset.py.
    .5 reassemble.py.
    .5 rlds.py.
    .5 rosbag\_ds.py.
    .5 \ldots .
    .4 utils/.
    .4 gui.py.
}
      \end{minipage}
    \end{minipage}
    \subcaption{}
    \label{fig:code_structure}
  \end{minipage}
    \caption{(a) Layout of the \textsc{ATLAS} annotation tool for a partially annotated episode. The top section shows multi-view camera visualizations, followed by a scrollable menu for selecting time-series signals and their corresponding plots. Below is the timeline for navigating the episode and the action annotations panel, where the last action is currently being edited. Control buttons are located at the bottom. (b) Action dialog for selecting labels and marking success. (c) Overview of the code structure.}

  \label{fig:gui_overview}
\end{figure*}

Users begin annotating a segment by pressing the \textit{Start Action} button or the corresponding keyboard shortcut. They can then navigate through the video using either the timeline or the assigned keyboard keys. Two navigation speeds are available, which can be configured in the configuration file. The faster speed allows rapid navigation of the video, while the slower speed enables precise placement of segmentation points. As the user navigates through the episode, a semi-transparent red overlay is dynamically displayed across the data plots to indicate the current annotation span. To end the segment, the user presses the same key used to start the annotation. An action dialog then appears, prompting the user to assign a semantic label to the annotated action, as shown in Figure~\ref{fig:action_dialog}. Additionally, a binary \textit{Success} flag can be toggled to mark the outcome of the action. Once confirmed, the new segment is added to the annotation panel.

\section{Code Architecture}
The software architecture of ATLAS is designed around the principles of modularity and extensibility, facilitating the integration of diverse robotic data standards. The system is split into a backend data-handling layer and a PyQt5-based frontend visualization layer, connected through a standardized abstraction interface.

\subsection{Backend - Data Abstraction Layer}
To achieve format-agnostic behavior, we implemented a template method design pattern~\cite{designpatterns1995} through an abstract base class, \texttt{DatasetBase}. This class defines a standardized interface for episode retrieval, metadata indexing, and data input/output operations. This architecture ensures high extensibility: by inheriting from \texttt{DatasetBase}, researchers can integrate new dataset formats, or even non-robotics multi-modal data, provided the implementation adheres to the defined abstraction.

Currently, we provide several implementations of \texttt{DatasetBase} that illustrate the flexibility of our design across different dataset formats:

\begin{itemize}
    \item \textbf{RLDS}~\cite{ramos2021rlds}: This format is based on TensorFlow Datasets (TFDS)~\cite{TFDS}. To balance responsiveness and memory usage, episodes are recursively converted into NumPy arrays at the episode level, minimizing latency without loading the full dataset into RAM.
    
    \item \textbf{Video}: Supports standard video formats (e.g., .mp4, .avi, .mkv) and uses OpenCV to decode video streams into NumPy arrays. An automatic detection mechanism handles three common storage layouts: single-file inputs, flat directories where each file corresponds to one episode, and multi-camera setups where subdirectories represent different camera views.
    
    \item \textbf{Frames}: Handles datasets in which videos are pre-extracted into frame directories. Folder structures are automatically detected using the same logic as in the \textit{Video} handler to ensure correct visualization.
    
    \item \textbf{Rosbags (ROS1 \& ROS2)}: Enables direct use of standard robotic logs (\texttt{.bag} and \texttt{.db3}) without requiring a ROS installation. We rely on the \texttt{rosbags} Python package~\cite{rosbags} and provide parsers for commonly used message types, including \texttt{JointState}, \texttt{WrenchStamped}, \texttt{PoseStamped}, \texttt{TwistStamped}, and \texttt{MultiArray}, which are converted into NumPy arrays.
\end{itemize}

In addition to these generic formats, we provide native support for the \textbf{REASSEMBLE} dataset~\cite{Sliwowski-RSS-25}, a contact-rich robotic manipulation dataset. In this dataset, each demonstration is stored as an HDF5 file. To reduce I/O latency, a dedicated background thread preloads adjacent data chunks into a ring buffer.
 
\subsection{Frontend - Data visualization and synchronization}
The challenge of synchronizing multi-modal data streams with varying sampling rates is addressed through nearest-neighbor temporal mapping. 
When the user interacts with the timeline slider, the system executes a binary search across the timestamp arrays of all enabled data streams. This ensures that the rendered video frames and the vertical cursor on the data plots are all synchronized.

The timestamp arrays on which this mapping operates are obtained differently depending on the dataset format. ROS bags record per-message timestamps for every topic, and REASSEMBLE provides individual timestamp arrays for each sensor modality, so the arrays can be read directly. RLDS instead stores all modalities in a shared step-indexed structure, which already temporally aligns observations, actions, and rewards by construction.
Video and frame-based datasets carry no explicit per-frame timestamps; in this case, the timeline is derived from a user-specified frame rate and all camera streams are assumed to be already time synchronized.

\subsection{Code Structure}
The structure of the code is illustrated in Figure~\ref{fig:code_structure}. The \texttt{annotations} folder stores the resulting JSON files containing the annotation data. The \texttt{configuration} directory contains dataset-specific configuration files. The \texttt{datasets} folder contains Python scripts implementing dataset-specific handlers. In particular, \texttt{dataset.py} implements the abstract template class \texttt{DatasetBase}, which users can inherit from to extend the GUI functionality for additional dataset formats. To add a new dataset, users should create a corresponding \texttt{my\_dataset.py} file within the \texttt{datasets} folder. The \texttt{utils} directory provides helper functions for parsing configuration files, operating on dictionaries, and managing different dataset formats. Finally, \texttt{gui.py} implements the frontend of the annotation tool.




\section{Experiments}

In our experiments, we aim to answer two questions:
\begin{itemize}
    \item [\textbf{Q1:}] How long does annotation take in ATLAS compared to other available
annotation tools?
    \item [\textbf{Q2:}] How do additional time-series visualizations influence annotation
alignment with expert annotations?
\end{itemize}

To quantify annotation alignment, we use two complementary metrics: a
continuous-time agreement indicator and the average boundary distance.
Let $T$ denote the total annotated duration and let
$y_a(t) \in \mathcal{C}$ be the piecewise-constant action label assigned by
annotator $a$ at time $t$.
The continuous-time agreement indicator between two annotators $a_1$ and $a_2$ is
defined as
\begin{equation}
A =
\frac{1}{T}
\int_0^T
\mathbf{1}\!\left[ y_{a_1}(t) = y_{a_2}(t) \right] dt.
\end{equation}
To capture boundary alignment, let $\mathcal{B}_{a}$ denote the set of temporal action boundaries annotated by annotator $a$, we define the asymmetric boundary distance
\begin{equation}
D_{a_1 \rightarrow a_2} =
\frac{1}{|\mathcal{B}_{a_1}|}
\sum_{b \in \mathcal{B}_{a_1}}
\min_{b' \in \mathcal{B}_{a_2}} |b - b'|.
\end{equation}
The symmetric boundary distance is then obtained by averaging both directions (annotator $a_1$ to $a_2$ and vice versa):
\begin{equation}
D_{\text{sym}} =
\frac{1}{2} \left(
D_{a_1 \rightarrow a_2} + D_{a_2 \rightarrow a_1}
\right).
\end{equation}

To evaluate annotation efficiency and consistency, we record four long-horizon
robot demonstrations of a gear assembly task using the NIST assembly task
board~\cite{Kimble_2020_NIST}. Specifically, we recorded three camera views alongside the robot's proprioceptive data, following the data acquisition protocol from REASSEMBLE~\cite{Sliwowski-RSS-25}.
One demonstration is used as a trial sequence to familiarize users with the
annotation tools, while the remaining three demonstrations, each spanning
\mbox{1--2} minutes, are used for evaluation.
We consider five action classes: \emph{grasp}, \emph{lift}, \emph{approach},
\emph{align}, and \emph{release}.
Expert annotations were obtained from two annotators with extensive experience on the NIST assembly task board. We computed their alignment score to be 99.6\% and the boundary distance to be 0.033s, indicating a high level of agreement. To generate a single ``ground truth'' annotation, the boundary points of each action were averaged across the two experts. These ground-truth annotations were then used to compare the annotations of other participants. In our experiments, the mean per-action annotation time is computed by dividing the total time required to annotate all three demonstrations by the total number of action segments. To compute the mean alignment score and boundary distance, for each annotator we concatenate the annotations from all recordings into a single long sequence to account for differences in video length. We report the mean and standard deviation of all metrics across annotators.

\subsection{Evaluation Conditions}
We evaluated four annotation conditions: ROSAnnotator~\cite{zhang2025rosannotator}, ELAN~\cite{wittenburg2006elan}, \textsc{ATLAS} with vision-only data, and \textsc{ATLAS} with vision plus time-series data. Each tool was configured in its most suitable setup for the annotation task, with the same set of five target action classes available for selection in all tools. The specific configurations were as follows:

\textbf{ROSAnnotator:} This tool only supports visualization of a single camera view, so we selected a representative view for display. Users could assign labels to the predefined action classes via a dropdown menu.

\textbf{ELAN:} Annotators could select actions from the same set of five classes using either the menu or keyboard shortcuts. Due to ELAN's limitations with mixed frame-rate playback, only two camera views were displayed, as the third view, recorded at a different frame rate, would become unsynchronized over time.

\textbf{ATLAS:} For vision-only annotation, three camera views were displayed. When including time-series data, plots of robot-measured forces and gripper width were shown by default. Annotators were free to adjust the displayed time-series data after the familiarization phase.

Twelve additional annotators with limited prior experience in robotic data annotation participated, with each assigned to a single tool variant. This setup was chosen to avoid learning effects; allowing participants to use multiple tools would make subsequent annotations trivial, as approximate action boundaries would already be known. The controls of the assigned tool were first explained, and the action definitions were presented. Each participant then had up to 10 minutes to familiarize themselves with the tool. Finally, participants were informed that they would now annotate three videos, that annotation accuracy should be prioritized, and that the annotation duration would be timed.

\begin{table}[t]
    \caption{Quantitative comparison of annotation performance across different tools. }
    \label{tab:Qual_results}
    \centering
    \begin{tabular}{C{2.5cm} C{1.5cm} C{1.5cm} C{1.5cm}}
        \toprule
        Condition & Average per-action annotation time (s) & Alignment score (\%) & Average boundary distance (s) \\
        \midrule
        ROSAnnotator~\cite{zhang2025rosannotator} & $26.9 \pm 6.1$ & $85.1 \pm 2.3$ & $1.59 \pm 0.25$\\
        ELAN~\cite{wittenburg2006elan} & $19.7 \pm 1.1$ & $96.7 \pm 0.8$ & $0.35 \pm 0.08$ \\
        Proposed \textsc{ATLAS} (vision-only) & $\mathbf{11.0 \pm 1.2}$ & $97.0 \pm 1.4$ & $0.31 \pm 0.15$ \\
        Proposed \textsc{ATLAS} (vision + time series) & $18.5 \pm 0.9$ & $\mathbf{99.4 \pm 0.1}$ & $\mathbf{0.06 \pm 0.01}$ \\
        \bottomrule
    \end{tabular}
\end{table}

\subsection{Discussion}

The results of our experiments are summarized in Table~\ref{tab:Qual_results}. Overall, the proposed \textsc{ATLAS} tool achieves lower average per-action annotation times compared to ROSAnnotator and ELAN. Both \textsc{ATLAS} and ELAN exhibit substantially faster annotation times than ROSAnnotator. We attribute this improvement to their keyboard-centric workflows, which reduce the need for mouse interactions and menu navigation, thereby accelerating the annotation process.

Comparing vision-only tools, \textsc{ATLAS} is on average 8.7~s faster per action than ELAN. We believe this difference is largely due to the complexity of ELAN’s keyboard shortcut system. ELAN is a highly flexible tool that supports a wide range of annotation tasks, resulting in a large set of shortcuts that annotators found difficult to remember, even after the familiarization phase which lasted up to 10 minutes. In contrast, \textsc{ATLAS} provides a smaller, customizable set of shortcuts tailored specifically to robotic action segmentation.

Interestingly, \textsc{ATLAS} with vision-only input is also 7.5~s faster than \textsc{ATLAS} with both vision and time-series data. We hypothesize that the additional sensor information increases the cognitive load during annotation, as users must analyze multiple data modalities simultaneously. This may slow down the annotation process; however, further experiments are required to validate this hypothesis.

Despite the increase in annotation time, incorporating robot proprioceptive data clearly improves annotation quality. As shown in Table~\ref{tab:Qual_results}, \textsc{ATLAS} with vision and time-series data achieves higher alignment scores and lower average boundary distances, with reduced standard deviation compared to vision-only tools. Proprioceptive signals provide clear cues for identifying action boundaries, leading to more consistent and accurate annotations.

Finally, ROSAnnotator achieves lower alignment scores mainly due to its limited temporal resolution. Its timeline is discretized to whole seconds, which makes it difficult to precisely align annotation boundaries with the video content. Although users can manually enter fractional start and end times, the coarse timeline visualization limits accuracy. This design choice reflects the original purpose of ROSAnnotator, which is intended for annotating social human–robot interactions, where sub-second precision is usually not required. In contrast, contact-rich robotic manipulation requires fine-grained temporal accuracy, as small timing differences can correspond to distinct physical events.


\textbf{Answering Q1}, \textsc{ATLAS} enables significantly faster annotation than existing tools, achieving the lowest average per-action time. It outperforms both ROSAnnotator and ELAN, which we attribute to a keyboard-centric design that reduces interaction overhead.

\textbf{Answering Q2}, although incorporating time-series visualizations increases annotation time, it leads to clear improvements in annotation quality. The additional proprioceptive cues result in higher alignment with expert annotations, and reduced variability across annotators, indicating more consistent and temporally precise annotations.

\section{Conclusions}
In this work, we propose \textsc{ATLAS}, an annotation tool designed specifically for robotic action segmentation. \textsc{ATLAS} natively supports five commonly used robotics dataset formats, including the Reinforcement Learning Dataset (RLDS) format~\cite{ramos2021rlds} employed by the Open X-Embodiment repository~\cite{o2024open}. Due to its modular software design, \textsc{ATLAS} can be easily extended to new datasets by implementing a simple dataset template interface. \textsc{ATLAS} follows a keyboard-centric interaction paradigm that minimizes reliance on mouse-based operations, leading to substantially reduced annotation times compared to existing tools. In our experiments, \textsc{ATLAS} with vision-only input achieves the fastest annotation speed, requiring $11.0 \pm 1.2$~s per action on average, compared to $19.7 \pm 1.1$~s for ELAN and $26.9 \pm 6.1$~s for ROSAnnotator. While incorporating time-series visualizations increases annotation time to $18.5 \pm 0.9$~s per action, it yields a substantial improvement in annotation quality, achieving an alignment score of $99.4 \pm 0.1\%$ and an average boundary distance of $0.06 \pm 0.01$~s. These results demonstrate that \textsc{ATLAS} enables both efficient and highly precise annotation of contact-rich robotic manipulation data.

Beyond its technical contribution, reducing the time and cost of annotating long-horizon robotic demonstrations carries broader societal and economic implications. High-quality annotated datasets are a key resource for modern robot learning, but their creation remains labor-intensive and expensive. More efficient annotation tools such as \textsc{ATLAS} lower this cost, making it easier and cheaper to build and extend datasets. At the same time, by reducing the human effort and time required for tedious labeling, tools like \textsc{ATLAS} help redirect annotators' time toward algorithm development.

\bibliographystyle{IEEEtran}
\bibliography{references}

\end{document}